\documentclass[journal,letterpaper]{IEEEtran}
\usepackage{xcolor}
\usepackage{cite}
\usepackage{amsmath,amssymb,amsfonts}
\usepackage{algorithmic}
\usepackage{algorithm}
\usepackage{graphicx}
\usepackage[percent]{overpic}
\usepackage{textcomp}
\usepackage{multirow}
\usepackage{tabularray}
\usepackage{nicematrix}
\usepackage{booktabs}
\usepackage{url}
\usepackage{stfloats}
\usepackage{balance}
\usepackage{needspace}
\renewcommand{\topfraction}{0.90}
\renewcommand{\dbltopfraction}{0.90}
\renewcommand{\textfraction}{0.08}
\renewcommand{\floatpagefraction}{0.90}
\renewcommand{\dblfloatpagefraction}{0.90}
\newlength\savewidth\newcommand\shline{\noalign{\global\savewidth\arrayrulewidth
  \global\arrayrulewidth 1.25pt}\hline\noalign{\global\arrayrulewidth\savewidth}}

\usepackage{comment}

\usepackage{ulem}

\def\BibTeX{{\rm B\kern-.05em{\sc i\kern-.025em b}\kern-.08em
    T\kern-.1667em\lower.7ex\hbox{E}\kern-.125emX}}
\makeatletter
\newcommand{\applybibcontrol}[1]{%
  \@bsphack\if@filesw\immediate\write\@auxout{\string\citation{#1}}\fi\@esphack}
\makeatother
\begin{document}
\applybibcontrol{TMIcontrol}
\title{UniField++: A Unified Agentic Framework with Loop Reasoning for MRI Field-Strength Enhancement}
\author{Yiyang~Lin, Chenhui~Wang, Tianliang~Yao, Pengli~Zhu, Shuaibo~Li, Zijun~Sun, Yun~Miao, Ang~Li, Yiming~Deng, and Yixuan~Yuan,~\IEEEmembership{Senior Member,~IEEE}
\thanks{This work was supported by the National Key Research and Development Program (2024YFE0216400) and Innovation and Technology Fund Mainland-Hong Kong Joint Funding Scheme (MHP/173/24). \textit{(Corresponding authors: Yixuan Yuan and Yiming Deng.)}}
\thanks{Yiyang Lin, Tianliang Yao, Pengli Zhu, Shuaibo Li, and Yixuan Yuan are with the Department of Electronic Engineering, The Chinese University of Hong Kong, Hong Kong SAR, China (e-mail: yylin247@163.com; yxyuan@ee.cuhk.edu.hk).}
\thanks{Chenhui Wang is with the Institute of Science and Technology for Brain-Inspired Intelligence, Fudan University, Shanghai 200433, China.}
\thanks{Zijun Sun is with the Institute of Diagnostic Radiology, University Hospital Zurich, University of Zurich, Zurich 8091, Switzerland.}
\thanks{Yun Miao is with the Theory Lab, Central Research Institute, 2012 Labs, Huawei Technologies Company, Ltd., Shanghai 200433, China.}
\thanks{Ang Li is with Yuanhua Orthopaedic Robotics (Shenzhen) Co., Ltd. (e-mail: ali@yuanhuatech.com).}
\thanks{Yiming Deng is with the Department of Neurology, Beijing Tiantan Hospital, Capital Medical University, Beijing 100070, China, and also with the China National Clinical Research Center for Neurological Diseases, Beijing 100070, China (e-mail: dengyiming@bjtth.org).}
}
\maketitle
\thispagestyle{plain}
\pagestyle{plain}

\begin{abstract}

Magnetic resonance imaging (MRI) field-strength enhancement seeks to combine the accessibility of low-field scanners with the image quality of high-field scanners.
Yet, existing methods remain prone to anatomical hallucinations: scarce paired data limit knowledge of field-strength-dependent tissue contrast and anatomical visibility, while uniform processing leaves difficult brain regions with insufficient reasoning and refinement. To address these challenges by better coupling field-aware learning with region-adaptive reasoning and refinement, we propose \textbf{UniField++}, a two-agent framework for unified multi-field enhancement, including 64mT-to-3T and 3T-to-7T. Unified learning exposes both the vision-language model (VLM)-based planner and the flow-matching-based enhancer to field-dependent image characteristics across tasks. Trained with radiologists' annotations of field-dependent hallucination-prone brain regions and corresponding tissue- and anatomy-specific correction plans, the planner predicts regional risk and formulates targeted restoration strategies. Guided by these plans, the enhancer performs loop reasoning along the flow-matching trajectory, repeatedly processing high-risk brain regions for additional reasoning and refinement. Together, multi-field unification and risk-guided reasoning support anatomically faithful enhancement across field strengths. Across the evaluated tasks, UniField++ improves PSNR by an average of 2.74\,dB and SSIM by 14.05 percentage points over comparison methods, with corresponding gains of 3.97\,dB and 13.92 percentage points over the ablation baseline. It also maintains enhancement quality on an external cohort with an unseen field strength and improves downstream whole-brain segmentation Macro Dice by 0.1457 on average over original low-field inputs. Code: \url{https://github.com/CUHK-AIM-Group/UniField_plusplus}.

\end{abstract}

\begin{IEEEkeywords}
Magnetic resonance imaging (MRI), Multi-agent systems, Unified enhancement, and Loop reasoning
\end{IEEEkeywords}
\vspace{-10pt}
\section{Introduction}

\label{sec:introduction}
Magnetic resonance imaging (MRI) is one of the most important non-invasive imaging modalities in modern clinical practice, with systems operating at magnetic field strengths ranging from ultra-low field, through the high field strengths commonly used in clinical settings, to ultra-high field. Specifically, portable low-field systems, such as 64mT scanners, are affordable and easy to deploy, but their low signal-to-noise ratio (SNR) limits spatial resolution and anatomical detail \cite{Arnold2023LowField,Kimberly2023LowField,Marques2019LowFieldPhysics}. Meanwhile, 1.5T and 3T systems are the mainstay of routine clinical MRI, providing higher SNR and clearer anatomical detail than ultra-low-field systems \cite{Arnold2023LowField}. Moreover, ultra-high-field 7T systems can further improve anatomical resolution and tissue contrast, but their wider adoption is constrained by greater financial and infrastructure requirements \cite{Radojewski2025SevenT}. 
To combine the affordability and deployment flexibility of low-field MRI with the excellent image quality of high-field MRI, computational field-strength enhancement aims to improve the anatomical detail and visual quality of low-field images toward those of high-field acquisitions.
\begin{figure}[t]
\vspace{-10pt}
    \centering
    \includegraphics[width=1.0\columnwidth]{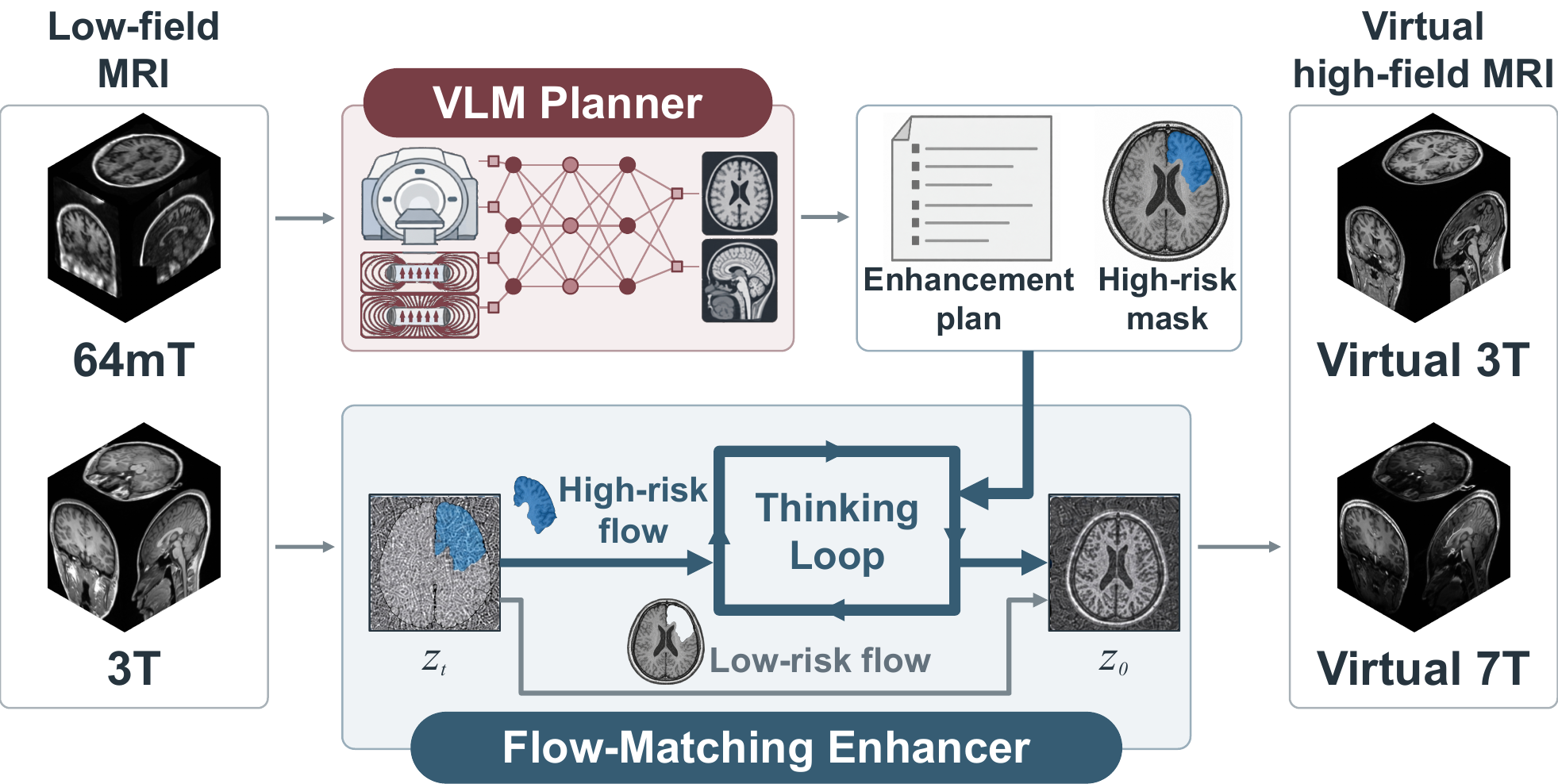}
    \caption{\textbf{Overview of UniField++.} Our framework unifies enhancement tasks across various field strengths. UniField++ primarily comprises a vision-language model (VLM)-based planner that simulates radiologists' planning process for enhancement and a flow-matching-based enhancer that executes the enhancement according to the plan.}
    \label{fig:overview}
\vspace{0pt}
\end{figure}

MRI field-strength enhancement has been studied across several field-strength pairs. For 64mT-to-3T enhancement, Islam et al. \cite{Islam2023LoHiResGAN} and Lucas et al. \cite{Lucas2025LowGAN} propose generative adversarial network (GAN)-based methods, while Baljer et al. \cite{Baljer2025MOUNet} propose a convolutional neural network (CNN)-based method using multiple scan orientations. For 1.5T-to-3T conversion, Jannat et al. \cite{Jannat2025Advancing} propose a transformer-enhanced GAN-based method. For 3T-to-7T synthesis, Qu et al. \cite{Qu2020WAT} and Cui et al. \cite{Cui2024SevenT} propose CNN-based methods, with the former integrating spatial and wavelet features, while Duan et al. \cite{Duan2024SynGAN} propose a GAN-based method. However, these approaches remain vulnerable to hallucinations, with two key limitations concerning the breadth of field-strength knowledge and the depth of regional reasoning. 
First, existing enhancement methods have limited knowledge of how field strength affects tissue contrast and the visibility of anatomical structures, often treating enhancement primarily as a low-level super-resolution task.
Limited paired datasets further constrain the acquisition of this knowledge: the KCL dataset contains 23 subjects for 64mT-to-3T enhancement \cite{vasa2025ultra}, while the UNC and BNU datasets contain only 10 and 20 subjects, respectively, for 3T-to-7T enhancement \cite{chen2023paired,chu2025paired}. Consequently, generated images may resemble high-field MRI while misrepresenting the patient's actual anatomy. For example, they may suppress a small lesion or introduce a lesion-like structure \cite{Cohen2018Hallucinations,Bhadra2021Hallucinations}.
Second, field-strength enhancement is not equally difficult across anatomical regions. While structures that are clearly visible in the low-field image may require relatively straightforward enhancement, poorly resolved or ambiguous regions require more careful interpretation and are more prone to hallucination~\cite{Tanno2021Uncertainty,Edupuganti2021Uncertainty,Bhadra2021Hallucinations}.
At the same time, existing methods generally use the same fixed processing depth across regions, leaving difficult regions without additional reasoning and refinement. Under this uniform allocation, even a model with broad prior knowledge may favor familiar structural patterns over patient-specific evidence. For example, it may reconstruct a poorly resolved sulcus as a tissue bridge between adjacent gyri. This motivates adaptive reasoning depth to examine and refine regions at greater hallucination risk.

Thus, we propose \textbf{UniField++}, a unified multi-agent framework for MRI field-strength enhancement, as illustrated in Fig.~\ref{fig:overview}. It comprises two cooperating agents: a vision-language model (VLM)-based planner initialized from pretrained Qwen 3.5 \cite{QwenTeam2026Qwen35} and a flow-matching-based enhancer initialized from pretrained FlashVSR \cite{Zhuang2026FlashVSR}, which provides restoration priors for our enhancement task. By bringing multiple field-strength transitions into a shared framework, UniField++ draws on a broader range of relationships between field strength, tissue contrast, and anatomical visibility than any single transition can provide. On the top of that, first, we incorporate radiologists' annotations of hallucinations and corresponding correction plans into the planner, enabling it to identify hallucination-prone regions and propose targeted corrections. Second, guided by the planner's risk assessments and correction plans, the enhancer performs loop reasoning along the flow-matching trajectory, repeatedly routing high-risk brain patches through a key processing stage for additional refinement. In this way, both agents support broader field-strength knowledge and adaptive reasoning depth to reduce hallucinations. 
Our contributions are as follows:

\begin{itemize}
    \item We propose \textbf{UniField++}, a unified multi-agent framework that decouples reasoning from enhancement through separate planner and enhancer, allowing each to focus on its task while reducing interference between the processes.

    \item We introduce a radiologist-informed VLM-based planner that learns field-strength knowledge through radiologist annotations, producing masks of high-risk regions and corresponding correction plans.

    \item We develop a flow-matching-based enhancer with dynamic reasoning that uses the masks and plans to allocate additional inference resources to high-risk regions through Loop Transformer blocks.

    \item UniField++ achieves average PSNR/SSIM gains of 2.74 dB/14.05 percentage points over competing methods and 3.97 dB/13.92 percentage points over the plain baseline in ablation experiments.
\end{itemize}

Compared with our preliminary framework, UniField \cite{Lin2026UniField}, published at MICCAI 2026, we introduce extensions in four major aspects. (1) We extend the single-network framework to a multi-agent framework, in which the planner focuses on reasoning to prevent hallucinations and the enhancer focuses on accurate generation. (2) We construct a VLM-based planner to prospectively identify regions prone to hallucination during field-strength enhancement and formulate explicit correction plans specifying restoration strategies. (3) We redesign the flow-matching enhancer to dynamically allocate its reasoning budget \cite{OpenAI2026GPT6} according to local hallucination risk predicted by the planner, allocating additional reasoning resources to high-risk regions through loop reasoning with a Loop Transformer. (4) These extensions consistently improve upon UniField across all five task--modality combinations, yielding average gains of 0.96\,dB in PSNR and 5.62 percentage points in SSIM.

\section{Related Works}
\subsection{MRI Field-Strength Enhancement}

MRI field-strength enhancement aims to combine the accessibility of low-field imaging with high-field image quality \cite{Arnold2023LowField,Kimberly2023LowField,Radojewski2025SevenT,PanHao_TaskAdaptive_MICCAISAT2026}. Existing studies investigate 64\,mT-to-3\,T \cite{Islam2023LoHiResGAN,Lucas2025LowGAN,Baljer2025MOUNet}, 1.5\,T-to-3\,T \cite{Liao2022Conversion,Jannat2025Advancing}, and 3\,T-to-7\,T enhancement \cite{Qu2020WAT,Duan2024SynGAN,Cui2024SevenT}. However, these studies largely treat these tasks independently despite shared requirements among different tasks, such as denoising and contrast recovery, that enable mutual improvement. In contrast, UniField \cite{Lin2026UniField} unifies multiple enhancement tasks to exploit this shared knowledge, improving enhancement performance. However, existing methods predominantly formulate enhancement as low-level super-resolution, with insufficient understanding of field-strength-dependent imaging characteristics, risking visually plausible outputs with tissue appearances inconsistent with the target field strength. In addition, limited paired data compound this insufficient understanding: the KCL dataset provides 23 subjects for 64mT-to-3T enhancement \cite{vasa2025ultra}, and the UNC and BNU datasets provide only 10 and 20 subjects for 3T-to-7T enhancement \cite{chen2023paired,chu2025paired}. In contrast, UniField++ learns field-dependent imaging characteristics from multiple enhancement tasks and radiologist annotations, using this knowledge to guide enhancement and improve image reliability.

\subsection{Agents for Generation}

Large language model (LLM)/VLM-based agents address complex tasks through coordinated reasoning, planning, and action. For instance, ReAct integrates reasoning traces and task-specific actions into an iterative loop \cite{Yao2023ReAct}, while HuggingGPT uses an LLM for task planning and model selection, with specialized models executing individual subtasks \cite{Shen2023HuggingGPT}. This paradigm has recently extended to visual generation. For instance, OmniHuman-1.5 employs multimodal large language model (MLLM)-based agents to interpret multimodal context and formulate action plans that guide a diffusion-based renderer \cite{Jiang2025OmniHuman15}. Similarly, Bernini separates an MLLM-based semantic planner from a Diffusion Transformer renderer, assigning reasoning and image synthesis to specialized components \cite{Liu2026Bernini}. Collectively, these studies motivate an agentic approach to MRI field-strength enhancement in which reasoning and planning precede image synthesis. This paradigm is well suited to the task, yet its use for explicit regional planning and risk-adaptive enhancement remains underexplored. Specifically, in low-field MRI, noise and limited resolution can obscure fine brain structures \cite{Arnold2022Sensitivity}, making their recovery dependent on reasoning about the remaining anatomical evidence. However, requiring a single model to handle both high-level planning and voxel-wise synthesis may create competing demands that compromise enhancement fidelity. In contrast, assigning these tasks to specialized agents is intended to reduce such interference. Accordingly, UniField++ uses a VLM-based planner for reasoning and planning based on field-strength knowledge, and a flow-matching-based enhancer for plan-guided generation.

\begin{figure*}[!tb]
\vspace{-10pt}
    \centering
    \includegraphics[width=0.95\textwidth]{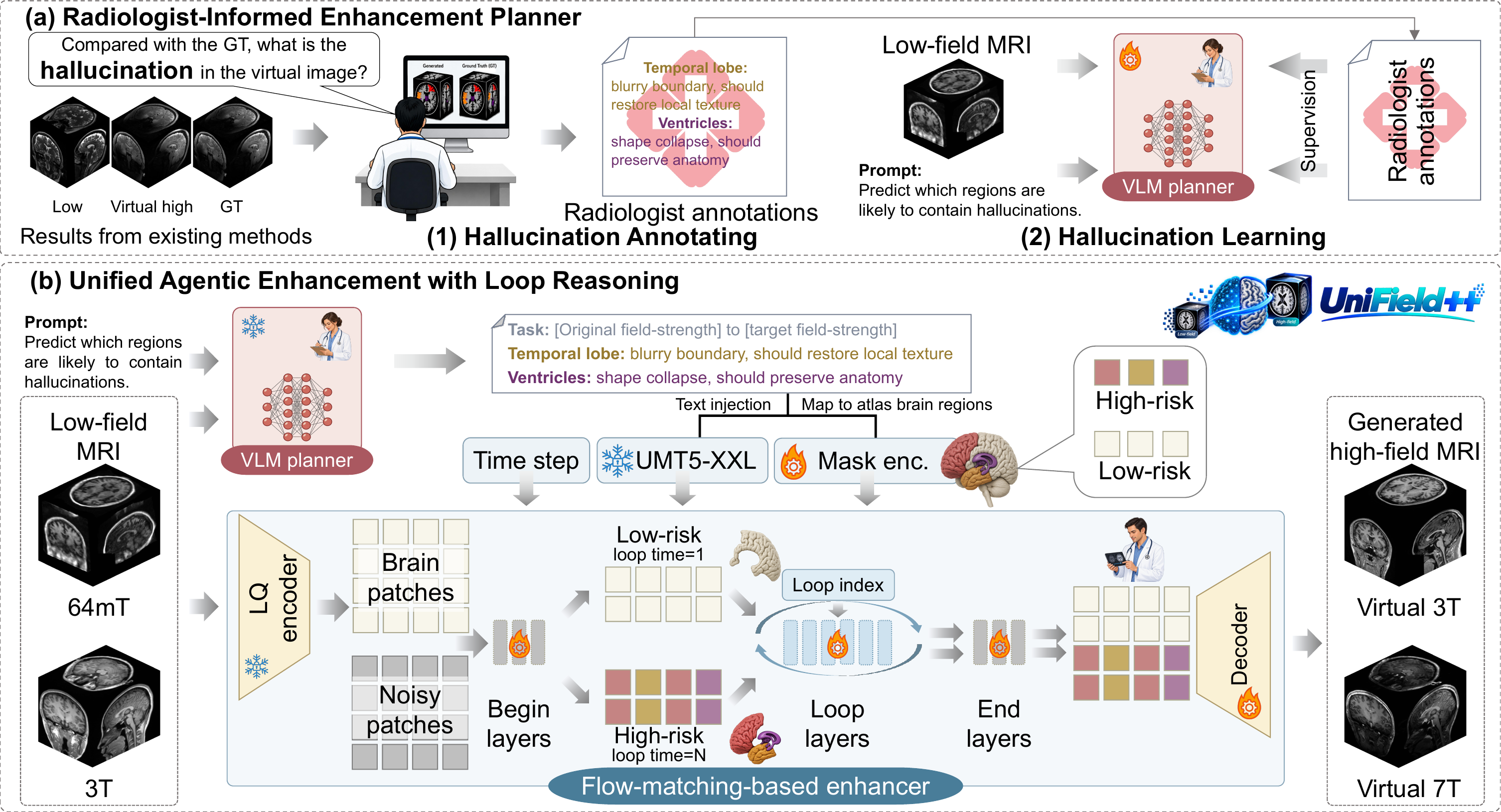}
    \caption{\textbf{Schematic diagram of UniField++.} Our framework mainly consists of two parts: (a) during construction of the \textbf{Radiologist-Informed Enhancement Planner (RIEP)}, (1) in the Hallucination Annotating stage, radiologists identify hallucinations in MRI field-strength enhancement and provide restoration plans; (2) in the hallucination learning stage, a VLM-based planner is fine-tuned to predict hallucinations and generate restoration plans; (b) during \textbf{Unified Agentic Enhancement with Loop Reasoning (UAELR)}, a flow-matching-based enhancer containing the loop transformer uses the plans provided by the VLM for MRI field-strength enhancement.}
    \label{fig:unifield_arch}
\vspace{-4pt}
\end{figure*}

\Needspace{5\baselineskip}
\section{Methodology}
\label{sec:method}
\subsection{Framework Overview}
\label{sec:method_overview}
In this paper, we propose \textbf{UniField++}, a unified multi-agent framework for MRI field-strength enhancement comprising a radiologist-informed VLM-based planner and a flow-matching-based enhancer. Firstly, as illustrated in Fig.~\ref{fig:unifield_arch}(a1), radiologists compare enhanced images with matched high-field references to identify hallucination-prone brain regions and provide targeted correction plans. Secondly, as shown in Fig.~\ref{fig:unifield_arch}(a2), the planner is fine-tuned with low-field images and these expert annotations so that it can anticipate hallucination risks and formulate enhancement plans directly from low-field images. Thirdly, as illustrated in Fig.~\ref{fig:unifield_arch}(b), the enhancer is conditioned on the predicted high-risk region masks and enhancement plans and performs iterative reasoning over hallucination-prone regions through Loop Transformer blocks. Then, the resulting high-risk patches are merged with low-risk patches and passed through the decoder to reconstruct a virtual high-field volume. The framework is formulated as
\begin{equation}
    \begin{aligned}
        (\boldsymbol{\mathcal{M}}, \boldsymbol{\mathcal{P}})
        &= \mathbf{P}\left(\mathbf{x}_{\mathrm{LF}}, \mathbf{b}_{\mathrm{HF}}\right), \\
        \hat{\mathbf{x}}_{\mathrm{HF}}
        &= \mathbf{E}\left(\mathbf{x}_{\mathrm{LF}}, \boldsymbol{\mathcal{M}}, \boldsymbol{\mathcal{P}}\right),
    \end{aligned}
\end{equation}
where $\mathbf{x}_{\mathrm{LF}}$ is the input low-field MRI, $\mathbf{b}_{\mathrm{HF}}$ is the target field strength, $\boldsymbol{\mathcal{M}}=\{\mathbf{M}_i\}_{i=1}^{J}$ contains the predicted region masks, and $\boldsymbol{\mathcal{P}}=\{\mathbf{p}_i\}_{i=1}^{J}$ contains their enhancement plans. Here $J$ is the number of selected regions, indexed by $i=1,\ldots,J$. The functions $\mathbf{P}(\cdot)$ and $\mathbf{E}(\cdot)$ denote the planner and enhancer, respectively, and $\hat{\mathbf{x}}_{\mathrm{HF}}$ is the enhanced target-field MRI.

During inference, the planner first predicts high-risk region masks and their enhancement plans. Then, the enhancer allocates additional Loop Transformer iterations to the high-risk patches and reconstructs a virtual high-field MRI volume.

\subsection{Radiologist-Informed Enhancement Planner}
\label{sec:radiologist_annotation}
\subsubsection{Hallucination Annotating}
As illustrated in Fig.~\ref{fig:unifield_arch}(a1), we collect structured annotations from radiologists to support hallucination anticipation and enhancement planning. Specifically, the radiologists compare the results of existing enhancement methods, including UniField and FlashVSR, with matched real high-field MRI. For each patient, we randomly sample several groups of images, with several slices in each group, and ask the radiologists to identify hallucination-prone brain regions, the corresponding enhancement issues, and appropriate enhancement plans.

Furthermore, to standardize annotations, we define three closed vocabularies: an anatomical region set $\boldsymbol{\mathcal{R}}$, an enhancement issue set $\boldsymbol{\mathcal{E}}$, and a plan vocabulary $\boldsymbol{\mathcal{V}}_{\mathrm{P}}$. Specifically, $\boldsymbol{\mathcal{R}}$ contains the frontal, parietal, temporal, occipital, and insular lobes, together with the ventricles and white matter. The issue set $\boldsymbol{\mathcal{E}}$ covers residual blur, over-smoothing, tissue-contrast mismatch, fluid-space mismatch, ringing and aliasing artifacts, noise mismatch, intensity inhomogeneity, local distortion, anatomical inconsistency, abnormal-signal errors, and clinically implausible over-correction. $\boldsymbol{\mathcal{V}}_{\mathrm{P}}$ contains edge de-ringing, detail-preserving denoising, tissue-contrast calibration, fine-detail recovery, intensity flattening, anti-aliasing, fluid-space refinement, artifact suppression, small-structure enhancement, geometry stabilization, abnormality-preserving enhancement, and conservative local refinement.

Formally, let $\boldsymbol{\mathcal{A}}(\mathbf{x}_{\mathrm{LF}}^{g})$ denote the annotation for one image group $g$, where $\mathbf{x}_{\mathrm{LF}}^{g}$ contains the source-field slices retrieved at the same $z$-axis locations as the virtual and real high-field slices assessed by the radiologists. Each annotation contains $J$ triplets, where $\mathbf{r}_i$, $\mathbf{e}_i$, and $\mathbf{p}_i$ specify the anatomical region, enhancement issue, and enhancement plan, respectively:
\begin{equation}
\begin{aligned}
    \boldsymbol{\mathcal{A}}(\mathbf{x}_{\mathrm{LF}}^{g})
    &= \left\{(\mathbf{r}_i,\mathbf{e}_i,\mathbf{p}_i)\right\}_{i=1}^{J},\\
    &\mathbf{r}_i \in \boldsymbol{\mathcal{R}},\quad
    \mathbf{e}_i \in \boldsymbol{\mathcal{E}},\quad
    \mathbf{p}_i \in \boldsymbol{\mathcal{V}}_{\mathrm{P}}.
\end{aligned}
\label{eq:doctor_annotation}
\end{equation}
As a result, these triplets describe where hallucinations arise, what form they take, and how the regions should be enhanced.

\subsubsection{Hallucination Learning}
\label{sec:planning_agent}
Building on these structured radiologist annotations, we use low-rank adaptation (LoRA) \cite{Hu2022LoRA} to fine-tune a Qwen3.5-9B-based planner \cite{QwenTeam2026Qwen35} to anticipate hallucinations during field-strength enhancement from low-field MRI, as shown by the pale-red module in Fig.~\ref{fig:unifield_arch}(a2). With its pretrained backbone frozen, it learns to predict region--issue--plan triplets and formulate enhancement plans.

Specifically, the source-field slice group $\mathbf{x}_{\mathrm{LF}}^{g}$ serves as the planner's visual input. The instruction tokens $\mathbf{q}_{\mathrm{inst}}$ encode the prompt \textit{Predict which regions are likely to contain hallucinations}, as shown in Fig.~\ref{fig:unifield_arch}(a2). The annotation is serialized into the target response $\mathbf{y}$.
In addition, let $\hat{\mathbf{y}}$ denote predicted token-probability distributions at supervised response positions, conditioned on the visual input, instruction, and preceding reference tokens. Finally, the planner is optimized with response-token cross-entropy (CE):
\begin{equation}
    \mathcal{L}_{\mathrm{plan}}
    = \operatorname{CE}\left(\hat{\mathbf{y}},\mathbf{y}\mid\mathbf{x}_{\mathrm{LF}}^{g},\mathbf{q}_{\mathrm{inst}}\right).
    \label{eq:planning_loss}
\end{equation}

\subsection{Unified Agentic Enhancement with Loop Reasoning}
\label{sec:enhancement_learning}
After Hallucination Learning, the frozen planner, shown in pale red in Fig.~\ref{fig:unifield_arch}(b), predicts $J$ hallucination-prone anatomical regions $\{\mathbf{r}_i\}_{i=1}^{J}$ and corresponding plans $\boldsymbol{\mathcal{P}}=\{\mathbf{p}_1,\ldots,\mathbf{p}_J\}$ that specify regional enhancement needs. To localize these regions in the patient's anatomy, we first register the MNI152NLin6Asym T1-weighted template \cite{Evans2012BrainTemplates} to the low-field MRI using sequential rigid, affine, and symmetric normalization (SyN) registration in Advanced Normalization Tools (ANTs) \cite{Avants2011ANTs}, with Mattes mutual information as the similarity metric. The resulting transformations are then applied to the Harvard--Oxford atlas labels \cite{Desikan2006Labeling} using label-preserving interpolation, providing coarse anatomical localization in the native low-field image space. We then merge the propagated atlas labels associated with each predicted region $\mathbf{r}_i$ to obtain the region-specific mask $\mathbf{M}_i$ and the mask set $\boldsymbol{\mathcal{M}}=\{\mathbf{M}_i\}_{i=1}^{J}$. 

To incorporate the predicted regional plans into the enhancement process, a frozen UMT5-XXL text encoder \cite{Chung2023UniMax} encodes the plans $\boldsymbol{\mathcal{P}}$ into semantic embeddings $\mathbf{e}_{\mathrm{plan}}$. Meanwhile, a multi-channel mask encoder with two bidirectional 3D convolutional stages and block-specific linear projections encodes the stacked masks $\boldsymbol{\mathcal{M}}$ into additive token embeddings $\mathbf{e}_{\mathrm{mask}}$. In addition, the frozen low-quality (LQ) encoder $\mathcal{E}_{\mathrm{LQ}}$ encodes the source-field MRI as $\mathbf{e}_{\mathrm{LF}}=\mathcal{E}_{\mathrm{LQ}}(\mathbf{x}_{\mathrm{LF}})$. These conditions guide the enhancer to generate virtual high-field MRI, as shown in the blue area of Fig.~\ref{fig:unifield_arch}(b). Specifically, the enhancer is a Loop Transformer-based flow-matching network, which will be introduced in the following paragraphs. 

\subsubsection{Routing Enhancement with Loop Transformers}
\label{sec:loop_transformer}
In fact, high-risk regions require in-depth reasoning to resolve ambiguous anatomical information. Otherwise, even with extensive field-strength knowledge, insufficient reasoning may merely assemble learned image characteristics into visually convincing but anatomically inconsistent structures. Actually, the Loop Transformer is well suited to this need: reusing the same Loop Layers increases the effective processing depth for high-risk regions through parameter sharing \cite{Dehghani2019Universal}, without introducing independently parameterized networks, thereby reducing model-storage and parameter-memory overhead.

Specifically, the 30-block enhancer backbone is organized into Begin, Loop, and End Layers. At the discrete flow-matching timestep $\tau$, the Begin Layers receive the current noisy latent $\mathbf{z}_{\tau}$, source embeddings $\mathbf{e}_{\mathrm{LF}}$, plan embeddings $\mathbf{e}_{\mathrm{plan}}$, mask embeddings $\mathbf{e}_{\mathrm{mask}}$, and timestep $\tau$:
\begin{equation}
    \mathbf{H}^{0}
    = \operatorname{Beg}\left(\mathbf{z}_{\tau},\mathbf{e}_{\mathrm{LF}},
    \mathbf{e}_{\mathrm{plan}},\mathbf{e}_{\mathrm{mask}},\tau\right),
    \label{eq:begin_layer}
\end{equation}
where $\mathbf{H}^{0}$ denotes features partitioned into low-risk patches $\mathbf{H}^{0}_{\mathrm{low}}$ and high-risk patches $\mathbf{H}^{0}_{\mathrm{high}}$ by the binary risk mask $\mathbf{M}$. This mask merges the regions in $\boldsymbol{\mathcal{M}}$ and is aligned to the feature or latent grid as needed. As shown in Fig.~\ref{fig:unifield_arch}(b), low-risk patches (off-white) pass through the Loop Layers once, whereas high-risk patches (salmon, yellow, and purple for different anatomical regions) undergo repeated refinement:
\begin{equation}
\begin{aligned}
    \mathbf{H}_{\mathrm{low}}
    &= \operatorname{Loop}\left(\mathbf{H}^{0}_{\mathrm{low}},1\right),\\
    \mathbf{H}^{(0)}_{\mathrm{high}} &= \mathbf{H}^{0}_{\mathrm{high}},\\
    \mathbf{H}^{(n)}_{\mathrm{high}}
    &= \operatorname{Loop}\left(\mathbf{H}^{(n-1)}_{\mathrm{high}},n\right).
\end{aligned}
\label{eq:risk_loop}
\end{equation}
Here $n=1,\ldots,N$ is the injected loop index, and $N$ is the number of iterations assigned to high-risk patches. Finally, the two subsets are restored to their original locations, re-aggregated, and transformed by the End Layers:
\begin{equation}
    \mathbf{v}_{\theta}
    = \operatorname{End}\left(\mathbf{H}_{\mathrm{low}}\oplus
    \mathbf{H}^{(N)}_{\mathrm{high}}\right),
    \label{eq:loop_merge}
\end{equation}
where $\mathbf{v}_{\theta}$ is the predicted velocity, $\theta$ denotes the backbone parameters, and $\oplus$ denotes spatial re-aggregation. This routing retains one pass for low-risk regions and allocates additional reasoning depth to regions more likely to hallucinate.

Together, the Begin, risk-guided Loop, and End Layers form the complete velocity-prediction backbone $\mathbf{F}_{\theta}$. Let $\mathbf{u}_{\tau}$ collect the inputs to the Begin Layers. The backbone is expressed as
\begin{equation}
 \mathbf{F}_{\theta}(\mathbf{u}_{\tau})
 = \bigl(\operatorname{End}\circ\operatorname{Loop}_{\mathbf{M}}^{(1,N)}
 \circ\operatorname{Beg}\bigr)(\mathbf{u}_{\tau}).
\label{eq:backbone}
\end{equation}
Here $\circ$ denotes function composition, and $\operatorname{Loop}_{\mathbf{M}}^{(1,N)}$ denotes the one-pass low-risk and $N$-pass high-risk processing in Eq.~\eqref{eq:risk_loop}, followed by spatial re-aggregation.

\subsubsection{Flow-Matching-Based Enhancer}
\label{sec:flow_matching_framework}
We formulate MRI field-strength enhancement within a flow-matching framework \cite{Lipman2023FlowMatching}, comprising the encoder, the Loop Transformer backbone introduced above, and the decoder. The encoder and decoder adopt the LQ encoder and tiny conditional (TC) decoder architectures of FlashVSR \cite{Zhuang2026FlashVSR}, respectively. Moreover, we initialize these components with weights from FlashVSR \cite{Zhuang2026FlashVSR}, a state-of-the-art video super-resolution model, for two reasons: (1) MRI volumes and videos share similar 3D data structures; and (2) video super-resolution and MRI field-strength enhancement both aim to suppress degradation and recover fine structural details. In addition, the encoder remains frozen, whereas the backbone and decoder are fine-tuned sequentially, as described below.

During training, we first define the forward process along a linear probability path from the clean target latent to Gaussian noise \cite{Liu2023RectifiedFlow}. Specifically, let $T=1000$, sample $\tau\in\{0,\ldots,T\}$, and set $t_{\tau}=\tau/T$. At $\tau=0$, the pretrained Wan 2.1 variational autoencoder (VAE) encoder $\mathcal{E}_{\mathrm{VAE}}$ \cite{Wan2025} encodes target-field MRI $\mathbf{x}_{\mathrm{HF}}$ as $\mathbf{z}_{0}=\mathcal{E}_{\mathrm{VAE}}(\mathbf{x}_{\mathrm{HF}})$. At $\tau=T$, we sample Gaussian noise $\mathbf{z}_{T}\sim\mathcal{N}(\mathbf{0},\mathbf{I})$. We then obtain the noisy latent $\mathbf{z}_{\tau}$ by linearly interpolating between $\mathbf{z}_{0}$ and $\mathbf{z}_{T}$ with weights $1-t_{\tau}$ and $t_{\tau}$, respectively. The target velocity $\mathbf{v}$ is $\mathbf{z}_{T}$ minus $\mathbf{z}_{0}$.

Then, the reverse process recovers a clean-latent estimate from Gaussian noise. The complete backbone $\mathbf{F}_{\theta}$ defined in Eq.~\eqref{eq:backbone} predicts the velocity $\mathbf{v}_{\theta}$ used for this process:
\begin{equation}
    \mathbf{v}_{\theta}
    = \mathbf{F}_{\theta}\!\left(\mathbf{z}_{\tau},t_{\tau}\mid
    \mathbf{e}_{\mathrm{LF}},\mathbf{e}_{\mathrm{plan}},\mathbf{e}_{\mathrm{mask}}\right).
    \label{eq:predicted_velocity}
\end{equation}
Finally, we train the flow-matching backbone with a risk-weighted $\ell_1$ objective, where $\lambda_{\mathrm{risk}}$ assigns greater supervision to high-risk regions predicted by the planner and represented by the merged mask $\mathbf{M}$. The same mask is used for both routing and loss weighting:
\begin{equation}
\begin{aligned}
    \mathcal{L}_{\mathrm{FM}}
    &= \mathbb{E}\!\left[
    \frac{\left\|\mathbf{W}\odot(\mathbf{v}_{\theta}-\mathbf{v})\right\|_{1}}
    {\|\mathbf{W}\|_{1}}\right],\\
    \mathbf{W} &= \mathbf{1}+(\lambda_{\mathrm{risk}}-1)\mathbf{M}.
\end{aligned}
\label{eq:flow_matching_loss}
\end{equation}
After training the flow-matching backbone, we freeze it and fine-tune the decoder, shown in pale yellow in Fig.~\ref{fig:unifield_arch}(b), in voxel space. Specifically, we sample $\mathbf{z}_{T}\sim\mathcal{N}(\mathbf{0},\mathbf{I})$ and evaluate the frozen backbone at $\tau=T=1000$, corresponding to normalized time $t_T=1$. Its velocity prediction yields a clean latent in one reverse step:
\begin{equation}
\begin{aligned}
    \mathbf{v}_{\theta}^{T}
    &= \mathbf{F}_{\theta}\!\left(\mathbf{z}_{T},1\mid
    \mathbf{e}_{\mathrm{LF}},\mathbf{e}_{\mathrm{plan}},\mathbf{e}_{\mathrm{mask}}\right),\\
    \hat{\mathbf{z}}_{0} &= \mathbf{z}_{T}-\mathbf{v}_{\theta}^{T}.
\end{aligned}
\label{eq:decoder_clean_latent}
\end{equation}
Then, the decoder $\mathcal{D}_{\mathrm{TC}}$ maps the clean-latent estimate $\hat{\mathbf{z}}_{0}$ to the reconstructed target-field MRI $\hat{\mathbf{x}}_{\mathrm{HF}}$.
Only the decoder is updated using an $\ell_1$ loss against paired target-field MRI $\mathbf{x}_{\mathrm{HF}}$, averaged over training pairs and sampled noise:
\begin{equation}
    \mathcal{L}_{\mathrm{dec}}
    = \mathbb{E}\!\left[
    \left\|\hat{\mathbf{x}}_{\mathrm{HF}}-\mathbf{x}_{\mathrm{HF}}\right\|_{1}\right].
    \label{eq:decoder_loss}
\end{equation}

\subsection{Objective Function and Inference}
\label{sec:training_inference}
UniField++ is trained in two stages: hallucination learning and UAELR. First, we train the RIEP using $\mathcal{L}_{\mathrm{plan}}$ in Eq.~\eqref{eq:planning_loss} to predict expert-defined region--issue--plan triplets. During UAELR, we train the backbone using $\mathcal{L}_{\mathrm{FM}}$ in Eq.~\eqref{eq:flow_matching_loss}, then freeze it and train the decoder using $\mathcal{L}_{\mathrm{dec}}$ in Eq.~\eqref{eq:decoder_loss}.

During inference, the planner first takes the source-field MRI $\mathbf{x}_{\mathrm{LF}}$ and predicts masks $\boldsymbol{\mathcal{M}}$ identifying hallucination-prone regions and plans $\boldsymbol{\mathcal{P}}$ specifying their enhancement needs. After that, the enhancer is conditioned on the encoded source image $\mathbf{e}_{\mathrm{LF}}$, plan embeddings $\mathbf{e}_{\mathrm{plan}}$, and mask embeddings $\mathbf{e}_{\mathrm{mask}}$ to produce virtual high-field volumes. Specifically, reverse sampling starts from $\hat{\mathbf{z}}_{T}\sim\mathcal{N}(\mathbf{0},\mathbf{I})$ along an $S$-step sampling schedule $T=\tau_0>\tau_1>\cdots>\tau_S=0$, with step index $s=0,\ldots,S-1$:
\begin{equation}
\begin{aligned}
    \hat{\mathbf{z}}_{\tau_{s+1}}
    &= \hat{\mathbf{z}}_{\tau_s}
    - \left(t_{\tau_s}-t_{\tau_{s+1}}\right)\\
    &\quad\times \mathbf{F}_{\theta}\!\left(
    \hat{\mathbf{z}}_{\tau_s},t_{\tau_s}\mid
    \mathbf{e}_{\mathrm{LF}},\mathbf{e}_{\mathrm{plan}},\mathbf{e}_{\mathrm{mask}}\right).
\end{aligned}
\label{eq:inference_flow}
\end{equation}
At the final step, $\tau_S=0$, so the resulting clean-latent estimate is $\hat{\mathbf{z}}_{0}=\hat{\mathbf{z}}_{\tau_S}$. After that, the decoder maps this estimate to the final target-field MRI $\hat{\mathbf{x}}_{\mathrm{HF}}$ using $\mathcal{D}_{\mathrm{TC}}$.

\setlength{\floatsep}{2pt plus 1pt minus 1pt}
\setlength{\dblfloatsep}{2pt plus 1pt minus 1pt}
\setlength{\textfloatsep}{4pt plus 1pt minus 1pt}
\setlength{\dbltextfloatsep}{4pt plus 1pt minus 1pt}
\setlength{\intextsep}{4pt plus 1pt minus 1pt}
\renewcommand{\topfraction}{0.88}
\renewcommand{\bottomfraction}{0.80}
\renewcommand{\dbltopfraction}{0.88}
\renewcommand{\textfraction}{0.08}
\renewcommand{\floatpagefraction}{0.95}
\renewcommand{\dblfloatpagefraction}{0.95}
\setcounter{topnumber}{3}
\setcounter{bottomnumber}{2}
\setcounter{dbltopnumber}{3}
\setcounter{totalnumber}{4}

\section{Experiments}
\NiceMatrixOptions{cell-space-limits=1.2pt}

\begin{table}[!tb]
\centering
\caption{\textbf{Dataset overview.} Three public datasets for 64mT-to-3T, two for 3T-to-7T, and a private 50mT-to-3T dataset are used.}
\label{Dataset}
\resizebox{\columnwidth}{!}{
\begin{NiceTabular}{c|c|c|c|ccc}[cell-space-limits=2.5pt]
\shline
\textbf{Task} & \textbf{Dataset} & \textbf{Modalities} & \textbf{Registered} & \textbf{Total} & \textbf{Train} & \textbf{Test} \\ 
\specialrule{\lightrulewidth}{0.8pt}{0.8pt}
\multirow{3}{*}{\textbf{64mT-to-3T}} & ULF-EnC \cite{Islam2023LoHiResGAN}, \cite{Dayarathna2024SynthesisReview,Dayarathna2025McCaD,Dayarathna2024ULFDiffusion} & T1, T2, FLAIR & Yes & 50 & 40 & 10 \\ 
                                     & Leiden \cite{vandenbroek2025paired} & T1, T2, FLAIR & No  & 10 & 8  & 2  \\ 
                                     & KCL \cite{vasa2025ultra}            & T1, T2        & No  & 23 & 18 & 5  \\ 
\hline
\multirow{2}{*}{\textbf{3T-to-7T}}   & UNC \cite{chen2023paired}           & T1, T2        & Yes & 10 & 8  & 2  \\ 
                                     & BNU \cite{chu2025paired}            & T1, T2        & No  & 20 & 16 & 4  \\ 
\hline
\textbf{50mT-to-3T} & BTH & T1, T2, FLAIR & Yes & 30 & 0 & 30 \\
\shline
\end{NiceTabular} 
}
\end{table}

\subsection{Dataset}

To train and evaluate UniField++, we employ five public datasets, each split into training and test sets at an approximate ratio of 4:1, as shown in Table~\ref{Dataset}. Moreover, we use a private dataset from Beijing Tiantan Hospital (BTH), comprising 30 patients with paired 50mT and 3T T1, T2, and FLAIR MRI, exclusively for external testing. We register unregistered low-/high-field pairs using nonlinear registration in ANTs \cite{Avants2011ANTs} with default parameters. Moreover, data preprocessing uses Medical Open Network for Artificial Intelligence (MONAI) \cite{Cardoso2022MONAI} for intensity normalization using the 0.5th--99.5th percentiles, 1\,mm resampling along the $z$-axis, and resizing to $256\times256\times160$. For hallucination learning in RIEP, radiologists randomly sample ten groups from each pair of virtual and real high-field volumes for annotation, with ten $z$-axis slices per group.

\begin{figure*}[!tb]
    \centering
    \includegraphics[width=0.82\textwidth]{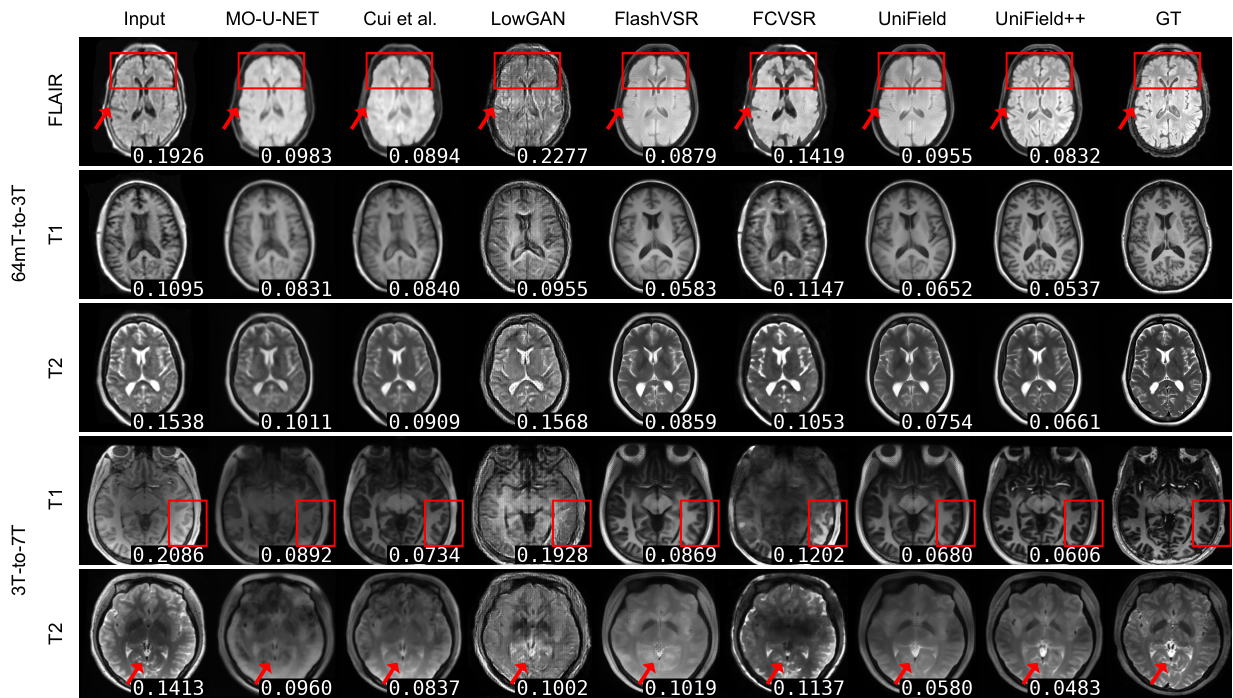}
\caption{\textbf{Qualitative comparison on the internal dataset for 64mT-to-3T and 3T-to-7T across multiple modalities.} The value in the lower-right corner of each image is the mean absolute error (MAE).} 
    \label{fig:comparative_experiments}
\end{figure*}

\subsection{Experimental Details}
For hallucination learning in RIEP, we use Adam \cite{Kingma2015Adam} with $\beta_1=0.5$ and $\beta_2=0.999$ and train for five epochs over all annotated samples. During the first 10\% of training iterations, the learning rate is linearly warmed up from $0$ to $2\times10^{-5}$, followed by cosine decay to $0$ over the remaining 90\% of iterations. This training is implemented in PyTorch \cite{Paszke2019PyTorch} and performed on one NVIDIA RTX A6000 GPU.

For UAELR, we adopt the same Adam optimizer and first train the generation backbone for 300 epochs, then independently train the decoder for 100 epochs. The learning rate starts from $5\times10^{-5}$ and is annealed to $1\times10^{-6}$ using a cosine schedule. We use PyTorch and four RTX A6000 GPUs, with three for enhancer training and one for planner inference.

We specify the main hyperparameters of the planner and enhancer. We set $J=3$ for the region--issue--plan triplets and corresponding region-specific masks and enhancement plans. For the planner, each sample contains ten source-field MRI slices evenly sampled along the $z$-axis across the volume, and Qwen3.5-9B is adapted using LoRA with rank $r=4$ and scaling factor $\alpha=16$. For the enhancer, both the Transformer backbone and the decoder are adapted using LoRA with rank $r=128$ and scaling factor $\alpha=128$.

\subsection{Comparative Experiments}

\begin{table*}[!tb]
  \centering
\caption{\textbf{64mT-to-3T comparison for T1, T2, and FLAIR.} NRMSE and SSIM are expressed in \%, PSNR is expressed in dB, and LPIPS is unitless. \textbf{Bold} and \underline{underlined} values indicate the best and second-best results.}
  \label{tab:64mt_to_3t_comparison}
  \setlength{\tabcolsep}{3.2pt}
  \renewcommand{\arraystretch}{1.0}
  \resizebox{0.98\textwidth}{!}{%
  \begin{NiceTabular}{l*{12}{c}}
    \toprule
    \multirow{2}{*}{Methods}
      & \multicolumn{4}{c}{T1}
      & \multicolumn{4}{c}{T2}
      & \multicolumn{4}{c}{FLAIR} \\
    \cmidrule(lr){2-5} \cmidrule(lr){6-9} \cmidrule(lr){10-13}
      & NRMSE $\downarrow$ & PSNR $\uparrow$ & SSIM $\uparrow$ & LPIPS $\downarrow$
      & NRMSE $\downarrow$ & PSNR $\uparrow$ & SSIM $\uparrow$ & LPIPS $\downarrow$
      & NRMSE $\downarrow$ & PSNR $\uparrow$ & SSIM $\uparrow$ & LPIPS $\downarrow$ \\
    \specialrule{\lightrulewidth}{0.8pt}{0.8pt}
    MO-U-NET (HBM'25)
      & 11.4 & 18.98 & 52.9 & 0.278
      & 9.5 & 20.58 & 59.2 & 0.233
      & 10.9 & 19.32 & 53.3 & 0.249 \\
    Cui et al. \cite{Cui2024SevenT} (MICCAI'24)
      & 11.3 & 19.05 & 67.9 & 0.255
      & 9.9 & 20.27 & 71.6 & 0.220
      & 10.9 & 19.31 & 65.6 & 0.238 \\
    LowGAN (Radiology'25)
      & 12.8 & 18.03 & 66.6 & 0.195
      & 11.4 & 18.98 & 65.8 & 0.195
      & 14.1 & 17.02 & 63.2 & \underline{0.195} \\
    FlashVSR (CVPR'26)
      & 12.1 & 18.41 & 63.4 & 0.298
      & 12.0 & 18.54 & 66.3 & 0.235
      & 15.1 & 16.61 & 60.6 & 0.300 \\
    FCVSR (TMM'26)
      & 14.1 & 17.16 & 61.7 & 0.258
      & 11.9 & 18.63 & 64.9 & 0.217
      & 16.6 & 15.67 & 55.6 & 0.258 \\
\specialrule{\lightrulewidth}{0.8pt}{0.8pt}
    UniField (MICCAI'26)
      & \underline{10.5} & \underline{19.75} & \underline{72.8} & \underline{0.185}
      & \underline{8.4} & \underline{21.57} & \underline{74.9} & \textbf{0.158}
      & \underline{9.5} & \underline{20.50} & \underline{71.5} & 0.200 \\
    UniField++ (Ours)
      & \textbf{9.6} & \textbf{20.53} & \textbf{77.6} & \textbf{0.171}
      & \textbf{7.8} & \textbf{22.19} & \textbf{78.7} & \underline{0.181}
      & \textbf{9.0} & \textbf{20.95} & \textbf{74.0} & \textbf{0.168} \\
\specialrule{\heavyrulewidth}{0.8pt}{0.8pt}
  \end{NiceTabular}
  }
\end{table*}

\begin{table}[!tb]
  \centering
\caption{\textbf{3T-to-7T comparison for T1 and T2.} \textbf{Bold} and \underline{underlined} values indicate the best and second-best results.}
  \label{tab:3t_to_7t_comparison}
  \scriptsize
  \setlength{\tabcolsep}{1.2pt}
  \renewcommand{\arraystretch}{1.0}
  \resizebox{\columnwidth}{!}{%
    \begin{NiceTabular}{c*{7}{c}}
      \toprule
      {\fontsize{6}{7}\selectfont\begin{tabular}[c]{@{}c@{}}Methods\end{tabular}}
        & {\fontsize{6}{7}\selectfont\begin{tabular}[c]{@{}c@{}}MO-U-NET\\HBM\\'25\end{tabular}}
        & {\fontsize{6}{7}\selectfont\begin{tabular}[c]{@{}c@{}}Cui et al.\,\cite{Cui2024SevenT}\\MICCAI\\'24\end{tabular}}
        & {\fontsize{6}{7}\selectfont\begin{tabular}[c]{@{}c@{}}LowGAN\\Radiology\\'25\end{tabular}}
        & {\fontsize{6}{7}\selectfont\begin{tabular}[c]{@{}c@{}}FlashVSR\\CVPR\\'26\end{tabular}}
        & {\fontsize{6}{7}\selectfont\begin{tabular}[c]{@{}c@{}}FCVSR\\TMM\\'26\end{tabular}}
        & {\fontsize{6}{7}\selectfont\begin{tabular}[c]{@{}c@{}}UniField\\MICCAI\\'26\end{tabular}}
        & {\fontsize{6}{7}\selectfont\begin{tabular}[c]{@{}c@{}}UniField++\\(Ours)\end{tabular}} \\
      \specialrule{\lightrulewidth}{0.8pt}{0.8pt}
      \multicolumn{8}{c}{T1} \\
      \specialrule{\lightrulewidth}{0.8pt}{0.8pt}
      NRMSE $\downarrow$ & 11.9 & 10.5 & 15.4 & 11.8 & 14.2 & \underline{9.9} & \textbf{8.2} \\
      PSNR $\uparrow$ & 18.51 & 19.66 & 16.26 & 18.63 & 16.97 & \underline{20.12} & \textbf{21.79} \\
      SSIM $\uparrow$ & 60.8 & \underline{66.8} & 54.0 & 63.0 & 56.8 & 66.1 & \textbf{77.1} \\
      LPIPS $\downarrow$ & 0.277 & 0.234 & 0.248 & 0.228 & 0.264 & \underline{0.207} & \textbf{0.198} \\
\specialrule{\lightrulewidth}{0.8pt}{0.8pt}
      \multicolumn{8}{c}{T2} \\
      \specialrule{\lightrulewidth}{0.8pt}{0.8pt}
      NRMSE $\downarrow$ & 8.6 & 8.1 & 10.1 & 9.5 & 11.5 & \underline{7.5} & \textbf{6.5} \\
      PSNR $\uparrow$ & 21.32 & 21.82 & 19.97 & 20.49 & 18.85 & \underline{22.55} & \textbf{23.85} \\
      SSIM $\uparrow$ & 59.7 & 65.0 & 56.0 & 58.6 & 54.8 & \underline{72.1} & \textbf{78.1} \\
      LPIPS $\downarrow$ & 0.274 & 0.245 & 0.254 & 0.239 & 0.254 & \textbf{0.176} & \underline{0.196} \\
\specialrule{\heavyrulewidth}{0.8pt}{0.8pt}
    \end{NiceTabular}%
  }
\end{table}

\subsubsection{Comparison on the Internal Dataset}

To demonstrate UniField++'s superior performance, we compare it with several recent MRI field-strength enhancement and video super-resolution methods. Firstly, MO-U-NET \cite{Baljer2025MOUNet} is a CNN-based method for 64mT-to-3T enhancement, LowGAN \cite{Lucas2025LowGAN} adopts a GAN-based framework for the same task, and Cui et al.'s CNN-based model \cite{Cui2024SevenT} is designed for 3T-to-7T enhancement. Additionally, UniField \cite{Lin2026UniField} serves as another competing method by unifying multiple field-strength enhancement tasks within a single framework. Moreover, we include two recent video super-resolution methods, FlashVSR \cite{Zhuang2026FlashVSR} and FCVSR \cite{Zhu2026FCVSR}, since video super-resolution and MRI field-strength enhancement share similar data structures and common restoration objectives such as noise suppression and detail recovery.

As shown in Fig.~\ref{fig:comparative_experiments}, MO-U-NET, Cui et al.'s model, and FlashVSR tend to over-smooth fine anatomical structures, as highlighted by the red arrows in the first and fifth rows, while LowGAN exhibits pronounced noise and checkerboard artifacts. In addition, FCVSR introduces evident local intensity distortions, producing excessively bright or dark regions, as indicated by the red boxes in the first and fourth rows. In contrast, UniField alleviates most of these artifacts, and UniField++ further improves the reconstruction by anticipating potential defect patterns before enhancement and providing corresponding restoration plans, directing more attention to defect-prone regions during enhancement. The quantitative results in Tables~\ref{tab:64mt_to_3t_comparison} and~\ref{tab:3t_to_7t_comparison} further demonstrate the effectiveness of UniField++. Compared with the six baselines across five tasks, UniField++ achieves average gains of 2.74 dB in PSNR and 14.05 percentage points in SSIM.

\begin{figure*}[!tb]
    \centering
    \includegraphics[width=0.82\textwidth]{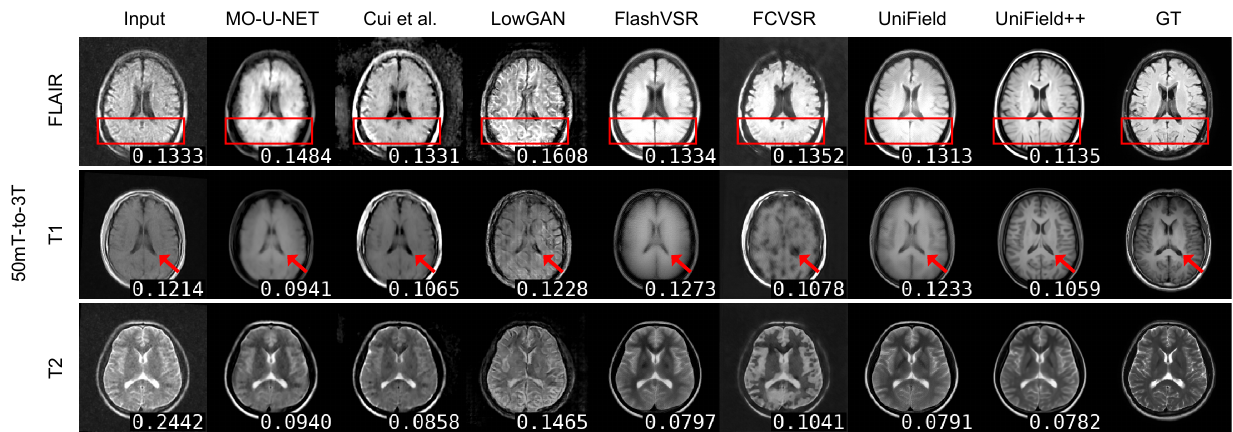}
\caption{\textbf{Qualitative comparison on the external dataset for the 50mT-to-3T task.}}
    \label{fig:generalizability}
\end{figure*}

\begin{table*}[!tb]
  \centering
  \caption{\textbf{External quantitative results on the 50mT-to-3T task.} \textbf{Bold} and \underline{underlined} values denote the best and second-best results, respectively; underlining is omitted when the best result is tied. }
  \label{tab:external_new3_all_steps}
  \setlength{\tabcolsep}{3.2pt}
  \renewcommand{\arraystretch}{1.0}
  \resizebox{0.98\textwidth}{!}{%
  \begin{NiceTabular}{l*{12}{c}}
    \toprule
    \multirow{2}{*}{Methods} & \multicolumn{4}{c}{T1} & \multicolumn{4}{c}{T2} & \multicolumn{4}{c}{FLAIR} \\
    \cmidrule(lr){2-5} \cmidrule(lr){6-9} \cmidrule(lr){10-13}
    & NRMSE~$\downarrow$ & PSNR~$\uparrow$ & SSIM~$\uparrow$ & LPIPS~$\downarrow$ & NRMSE~$\downarrow$ & PSNR~$\uparrow$ & SSIM~$\uparrow$ & LPIPS~$\downarrow$ & NRMSE~$\downarrow$ & PSNR~$\uparrow$ & SSIM~$\uparrow$ & LPIPS~$\downarrow$ \\
    \specialrule{\lightrulewidth}{0.8pt}{0.8pt}
    MO-U-NET (HBM'25) & 9.7 & 20.32 & 76.6 & 0.212 & 9.0 & 21.11 & 79.6 & 0.184 & 12.5 & 18.10 & 74.2 & 0.227 \\
    Cui et al. \cite{Cui2024SevenT} (MICCAI'24) & 10.1 & 20.05 & \underline{77.2} & \textbf{0.174} & 9.2 & 20.87 & 79.3 & 0.174 & 13.9 & 17.21 & 74.2 & 0.215 \\
    LowGAN (Radiology'25) & 10.0 & 20.12 & 73.6 & \textbf{0.174} & 11.5 & 18.84 & 72.6 & 0.180 & 13.7 & 17.28 & 70.8 & \underline{0.191} \\
    FlashVSR (CVPR'26) & \textbf{9.2} & \textbf{20.88} & 75.7 & 0.249 & 8.8 & 21.32 & 78.1 & 0.176 & 12.2 & 18.34 & 74.6 & 0.232 \\
    FCVSR (TMM'26) & 10.5 & 19.75 & 75.7 & 0.190 & 9.3 & 20.77 & 78.8 & \textbf{0.158} & \textbf{11.7} & \textbf{18.69} & 75.1 & \textbf{0.179} \\
    \specialrule{\lightrulewidth}{0.8pt}{0.8pt}
    UniField (MICCAI'26) & \textbf{9.2} & \underline{20.87} & \textbf{77.3} & 0.196 & 8.9 & 21.22 & 78.5 & \underline{0.168} & 12.2 & 18.29 & 75.4 & 0.222 \\
    UniField++ (1 step) & \textbf{9.2} & 20.80 & \underline{77.2} & 0.195 & \textbf{8.4} & \underline{21.69} & \underline{80.0} & 0.190 & \underline{12.1} & \underline{18.36} & \underline{76.1} & 0.239 \\
    UniField++ (2 steps) & 9.3 & 20.72 & 76.7 & 0.184 & \textbf{8.4} & \textbf{21.70} & \textbf{80.1} & 0.185 & 12.2 & 18.32 & \textbf{76.2} & 0.233 \\
    UniField++ (10 steps) & 9.4 & 20.60 & 75.7 & \textbf{0.174} & 8.7 & 21.46 & 79.3 & 0.173 & 12.7 & 17.96 & 75.1 & 0.220 \\
    \specialrule{\heavyrulewidth}{0.8pt}{0.8pt}
  \end{NiceTabular}
  }
\end{table*}

\subsubsection{Comparison on the External Dataset}

To demonstrate the generalization capabilities of UniField++, we evaluate UniField++ with all six competing methods on the BTH external test cohort, consisting of 30 patients with paired 50mT and 3T MRI in T1, T2, and FLAIR modalities. Notably, images acquired at 50 mT are not included in the training set, and thus the 50mT-to-3T task can effectively measure the generalization capabilities of all the methods.

As shown in Fig.~\ref{fig:generalizability}, the images generated by UniField++ closely resemble the corresponding high-field MRI across different modalities. In comparison, MO-U-NET, Cui et al.'s model, and FlashVSR lose substantial anatomical details, while LowGAN and FCVSR introduce hallucinated structures, as highlighted by the red boxes in the first row and the red arrows in the second row. Table~\ref{tab:external_new3_all_steps} shows that UniField++ ranks best or second best on most metrics, supporting generalization to the unseen source field strength.

\begin{figure*}[!tb]
    \centering
    \includegraphics[width=0.74\textwidth]{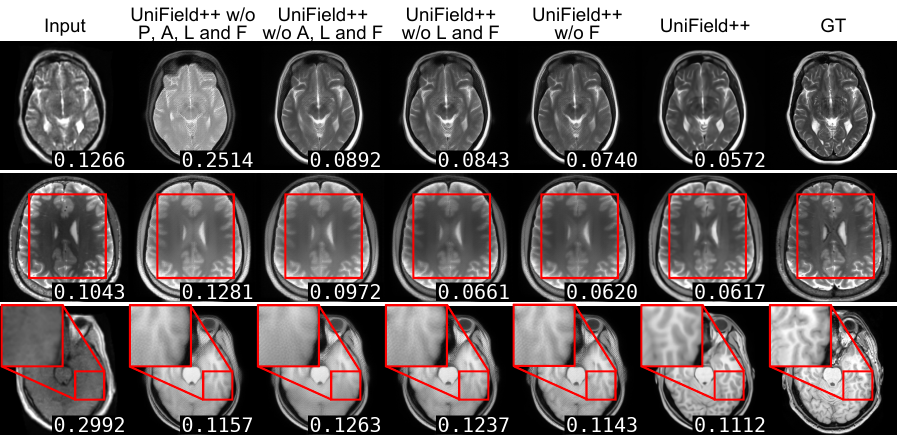}
    \caption{\textbf{Qualitative ablation of components in UAELR.} Column labels: P, plan; A, atlas; L, loop; F, decoder fine-tuning. Red boxes and zoom-ins highlight the changes introduced by each component.}
    \label{fig:ablation_experiments}
\end{figure*}

\subsection{Ablation Experiments}

\subsubsection{Ablation on Hallucination Learning in RIEP}
To evaluate hallucination learning in RIEP, we use Region F1, Issue F1, and Plan F1 to measure set-based overlap between predictions and references, and the Semantic score to measure the similarity between VLM outputs and radiologist annotations by averaging sequence similarity, token F1, ROUGE-L F1, character-trigram F1, and bag-of-words cosine similarity. As shown in Table~\ref{tab:medical_vlm_finetune}, hallucination learning improves all metrics, with the largest gain of 23.97 percentage points in Issue F1, demonstrating a substantially improved ability to anticipate hallucination risks during field-strength enhancement.

\begin{table}[!tb]
  \centering
  \caption{\textbf{Hallucination learning (HL) ablation.}
  }
  \label{tab:medical_vlm_finetune}
  \setlength{\tabcolsep}{5pt}
  \renewcommand{\arraystretch}{1.0}
  \begin{NiceTabular}{ccccc}
    \toprule
    Setting & Region F1 $\uparrow$ & Issue F1 $\uparrow$ & Plan F1 $\uparrow$ & Semantic $\uparrow$ \\
    \specialrule{\lightrulewidth}{0.8pt}{0.8pt}
    W/o HL  & 0.6546 & 0.3986 & 0.6833 & 0.3548 \\
    With HL & \textbf{0.8511} & \textbf{0.6383} & \textbf{0.8313} & \textbf{0.5213} \\
\specialrule{\heavyrulewidth}{0.8pt}{0.8pt}
  \end{NiceTabular}
\end{table}

\begin{table*}[!tb]
  \centering
  \caption{\textbf{Ablation results on the 64mT-to-3T task.} NRMSE and SSIM are expressed in \%, PSNR is expressed in dB, and LPIPS is unitless. \textbf{Bold} and \underline{underlined} values indicate the best and second-best results, respectively; underlining is omitted when the best result is tied. $\lambda_{\mathrm{risk}}$ is the risk weight in Eq.~\eqref{eq:flow_matching_loss}; the loop block numbers specify the blocks used for cyclic inference in the 30-block backbone.}
  \vspace{-3pt}
  \label{tab:64mt_to_3t_ablation}
  \setlength{\tabcolsep}{3.0pt}
  \renewcommand{\arraystretch}{0.96}
  \resizebox{0.98\textwidth}{!}{%
  \begin{NiceTabular}{cccc|cc*{12}{c}}
    \toprule
    \multicolumn{2}{c}{Planner} & \multicolumn{2}{c|}{Enhancer} & \multirow{2}{*}{\begin{tabular}[c]{@{}c@{}}Hyper-\\parameters\end{tabular}} & \multirow{2}{*}{Setting} & \multicolumn{4}{c}{T1} & \multicolumn{4}{c}{T2} & \multicolumn{4}{c}{FLAIR} \\
    \cmidrule(lr){1-2} \cmidrule(lr){3-4} \cmidrule(lr){7-10} \cmidrule(lr){11-14} \cmidrule(lr){15-18}
    Plan & Atlas & Loop & \begin{tabular}[c]{@{}c@{}}Fine-tune \\ decoder\end{tabular} & & & NRMSE $\downarrow$ & PSNR $\uparrow$ & SSIM $\uparrow$ & LPIPS $\downarrow$ & NRMSE $\downarrow$ & PSNR $\uparrow$ & SSIM $\uparrow$ & LPIPS $\downarrow$ & NRMSE $\downarrow$ & PSNR $\uparrow$ & SSIM $\uparrow$ & LPIPS $\downarrow$ \\
    \specialrule{\lightrulewidth}{0.8pt}{0.8pt}
    $\times$ & $\times$ & $\times$ & $\times$ & -- & -- & 13.1 & 17.77 & 68.1 & 0.252 & 14.7 & 16.82 & 60.9 & 0.268 & 14.4 & 16.99 & 63.5 & 0.282 \\
    \specialrule{\lightrulewidth}{0.8pt}{0.8pt}
    $\checkmark$ & $\times$ & $\times$ & $\times$ & -- & -- & 10.9 & 19.38 & 71.2 & 0.194 & 8.7 & 21.26 & 71.6 & 0.171 & 10.0 & 20.05 & 67.0 & 0.212 \\
    \specialrule{\lightrulewidth}{0.8pt}{0.8pt}
    \multirow{3}{*}{$\checkmark$} & \multirow{3}{*}{$\checkmark$} & \multirow{3}{*}{$\times$} & \multirow{3}{*}{$\times$} & \Block{3-1}{$\lambda_{\mathrm{risk}}$} & 1 & 10.9 & 19.40 & 72.9 & 0.190 & 9.0 & 21.03 & 74.1 & 0.163 & 10.1 & 19.95 & 69.4 & 0.206 \\
     &  &  &  &  & 5 & 10.7 & 19.47 & 72.7 & 0.184 & 8.8 & 21.21 & 74.1 & 0.164 & 9.6 & 20.40 & 70.9 & 0.197 \\
     &  &  &  &  & 10 & 10.9 & 19.31 & 70.0 & 0.193 & 8.7 & 21.25 & 73.1 & 0.171 & 10.8 & 19.45 & 66.9 & 0.210 \\
    \specialrule{\lightrulewidth}{0.8pt}{0.8pt}
    \multirow{6}{*}{$\checkmark$} & \multirow{6}{*}{$\checkmark$} & \multirow{6}{*}{$\checkmark$} & \multirow{6}{*}{$\times$} & \Block{3-1}{\shortstack{Loop block\\numbers}} & 12--16 & 10.6 & 19.69 & 74.0 & 0.176 & 8.6 & 21.42 & 74.7 & 0.160 & 9.5 & 20.47 & 71.4 & 0.188 \\
     &  &  &  &  & 10--19 & 10.4 & 19.80 & 74.4 & 0.173 & 8.5 & 21.48 & 75.0 & \textbf{0.156} & 9.4 & 20.53 & 71.7 & 0.190 \\
     &  &  &  &  & 5--24 & 10.3 & 19.91 & 72.0 & 0.193 & 8.6 & 21.41 & 72.9 & 0.168 & 9.6 & 20.40 & 68.7 & 0.209 \\
    \cmidrule(lr){5-18}
     &  &  &  & \Block{3-1}{Loop times} & 1 & 10.4 & 19.80 & 74.4 & 0.173 & 8.5 & 21.48 & 75.0 & \textbf{0.156} & 9.4 & 20.53 & 71.7 & 0.190 \\
     &  &  &  &  & 2 & 10.4 & 19.81 & 72.9 & 0.184 & 8.7 & 21.26 & 74.0 & 0.169 & 9.7 & 20.33 & 70.3 & 0.196 \\
     &  &  &  &  & 5 & 10.1 & 20.00 & 72.9 & 0.185 & 8.7 & 21.27 & 73.6 & 0.166 & 9.8 & 20.17 & 69.6 & 0.203 \\
    \specialrule{\lightrulewidth}{0.8pt}{0.8pt}
    \multirow{6}{*}{$\checkmark$} & \multirow{6}{*}{$\checkmark$} & \multirow{6}{*}{$\checkmark$} & \multirow{6}{*}{$\checkmark$} & \Block{6-1}{Sampling steps} & 1 & \textbf{9.2} & \textbf{20.95} & \textbf{78.3} & 0.200 & \underline{7.7} & \underline{22.38} & \underline{78.7} & 0.197 & \textbf{8.6} & \textbf{21.34} & \underline{74.9} & 0.202 \\
     &  &  &  &  & 2 & \underline{9.3} & \underline{20.86} & \textbf{78.3} & 0.188 & \textbf{7.6} & \textbf{22.48} & \textbf{79.4} & 0.186 & \textbf{8.6} & \textbf{21.34} & \textbf{75.0} & 0.188 \\
     &  &  &  &  & 5 & 9.6 & 20.53 & 77.6 & 0.171 & 7.8 & 22.19 & \underline{78.7} & 0.181 & 9.0 & 20.95 & 74.0 & 0.168 \\
     &  &  &  &  & 10 & 9.9 & 20.27 & 77.1 & 0.164 & 8.1 & 21.85 & 78.1 & 0.179 & 9.3 & 20.69 & 73.4 & 0.161 \\
     &  &  &  &  & 20 & 10.1 & 20.05 & 76.6 & \underline{0.161} & 8.3 & 21.66 & 77.8 & 0.178 & 9.4 & 20.59 & 73.0 & \underline{0.160} \\
     &  &  &  &  & 50 & 10.3 & 19.87 & 76.3 & \textbf{0.158} & 8.5 & 21.49 & 77.6 & 0.171 & 9.6 & 20.38 & 72.7 & \textbf{0.150} \\
\specialrule{\heavyrulewidth}{0.8pt}{0.8pt}
  \end{NiceTabular}%
  }
\vspace{-5pt}
\end{table*}

\begin{table*}[!tb]
  \centering
  \caption{\textbf{Ablation results on the 3T-to-7T task.} \textbf{Bold} and \underline{underlined} values indicate the best and second-best results, respectively; underlining is omitted when the best result is tied. $\lambda_{\mathrm{risk}}$ is the risk weight in Eq.~\eqref{eq:flow_matching_loss}; the loop block numbers specify the blocks used for cyclic inference in the 30-block backbone.}
  \vspace{-3pt}
  \label{tab:3t_to_7t_ablation}
  \fontsize{10}{12}\selectfont
  \setlength{\tabcolsep}{2.97pt}
  \renewcommand{\arraystretch}{0.88}
  \scalebox{0.675}{%
  \begin{NiceTabular}{cccc|cc*{8}{c}}
    \toprule
    \multicolumn{2}{c}{Planner} & \multicolumn{2}{c|}{Enhancer} & \multirow{2}{*}{\begin{tabular}[c]{@{}c@{}}Hyper-\\parameters\end{tabular}} & \multirow{2}{*}{Setting} & \multicolumn{4}{c}{T1} & \multicolumn{4}{c}{T2} \\
    \cmidrule(lr){1-2} \cmidrule(lr){3-4} \cmidrule(lr){7-10} \cmidrule(lr){11-14}
    Plan & Atlas & Loop & \begin{tabular}[c]{@{}c@{}}Fine-tune \\ decoder\end{tabular} & & & NRMSE $\downarrow$ & PSNR $\uparrow$ & SSIM $\uparrow$ & LPIPS $\downarrow$ & NRMSE $\downarrow$ & PSNR $\uparrow$ & SSIM $\uparrow$ & LPIPS $\downarrow$ \\
    \specialrule{\lightrulewidth}{0.8pt}{0.8pt}
    $\times$ & $\times$ & $\times$ & $\times$ & -- & -- & 12.4 & 18.22 & 60.7 & 0.261 & 10.5 & 19.66 & 62.7 & 0.231 \\
    \specialrule{\lightrulewidth}{0.8pt}{0.8pt}
    $\checkmark$ & $\times$ & $\times$ & $\times$ & -- & -- & 11.3 & 18.96 & 64.4 & 0.210 & 8.4 & 21.61 & 65.3 & 0.188 \\
    \specialrule{\lightrulewidth}{0.8pt}{0.8pt}
    \multirow{3}{*}{$\checkmark$} & \multirow{3}{*}{$\checkmark$} & \multirow{3}{*}{$\times$} & \multirow{3}{*}{$\times$} & \Block{3-1}{$\lambda_{\mathrm{risk}}$} & 1 & 10.0 & 20.05 & 67.9 & 0.206 & 8.3 & 21.69 & 68.5 & 0.178 \\
     &  &  &  &  & 5 & 10.2 & 19.92 & 67.0 & 0.198 & 7.5 & 22.48 & 71.9 & 0.173 \\
     &  &  &  &  & 10 & 9.6 & 20.49 & 68.7 & 0.200 & 8.1 & 21.87 & 70.7 & 0.178 \\
    \specialrule{\lightrulewidth}{0.8pt}{0.8pt}
    \multirow{6}{*}{$\checkmark$} & \multirow{6}{*}{$\checkmark$} & \multirow{6}{*}{$\checkmark$} & \multirow{6}{*}{$\times$} & \Block{3-1}{\shortstack{Loop block\\numbers}} & 12--16 & 10.0 & 20.06 & 68.2 & \underline{0.189} & 8.3 & 21.67 & 69.6 & 0.176 \\
     &  &  &  &  & 10--19 & 10.0 & 20.06 & 68.4 & 0.199 & 8.1 & 21.93 & 69.9 & 0.175 \\
     &  &  &  &  & 5--24 & 9.8 & 20.25 & 68.1 & 0.203 & 7.6 & 22.42 & 70.3 & 0.174 \\
    \cmidrule(lr){5-14}
     &  &  &  & \Block{3-1}{Loop times} & 1 & 10.0 & 20.06 & 68.4 & 0.199 & 8.1 & 21.93 & 69.9 & 0.175 \\
     &  &  &  &  & 2 & 9.8 & 20.26 & 69.1 & \underline{0.189} & 7.7 & 22.26 & 72.0 & \textbf{0.170} \\
     &  &  &  &  & 5 & 10.5 & 19.64 & 66.9 & 0.197 & 7.9 & 22.04 & 69.9 & \textbf{0.170} \\
\specialrule{\lightrulewidth}{0.8pt}{0.8pt}
    \multirow{6}{*}{$\checkmark$} & \multirow{6}{*}{$\checkmark$} & \multirow{6}{*}{$\checkmark$} & \multirow{6}{*}{$\checkmark$} & \Block{6-1}{Sampling steps} & 1 & \textbf{8.1} & \textbf{22.00} & \underline{77.6} & 0.229 & \textbf{6.3} & \underline{24.04} & \underline{78.2} & 0.217 \\
     &  &  &  &  & 2 & \textbf{8.1} & \underline{21.94} & \textbf{77.7} & 0.216 & \textbf{6.3} & \textbf{24.08} & \textbf{78.7} & 0.208 \\
     &  &  &  &  & 5 & 8.2 & 21.79 & 77.1 & 0.198 & 6.5 & 23.85 & 78.1 & 0.196 \\
     &  &  &  &  & 10 & 8.5 & 21.51 & 76.1 & \underline{0.189} & 6.6 & 23.61 & 77.5 & 0.190 \\
     &  &  &  &  & 20 & 8.7 & 21.34 & 75.2 & \textbf{0.184} & 6.7 & 23.48 & 77.6 & 0.182 \\
     &  &  &  &  & 50 & 8.7 & 21.28 & 74.4 & 0.193 & 6.8 & 23.34 & 78.1 & 0.179 \\
\specialrule{\heavyrulewidth}{0.8pt}{0.8pt}
  \end{NiceTabular}%
  }
\vspace{-5pt}
\end{table*}

\subsubsection{Ablation on components in UAELR}
To validate the effectiveness of each component in UniField++, we conduct cumulative ablation experiments in this section. Within the planner, firstly, the VLM formulates restoration plans that improve overall intensity fidelity and noise suppression when injected into the enhancer, as shown in the first row (second and third columns) of Fig.~\ref{fig:ablation_experiments}. This is because the VLM planner has been exposed to a large number of generation-failure cases during training and can anticipate and avoid these common degradation patterns, such as intensity errors and residual noise. 
Secondly, incorporating the atlas-based hallucination risk mask makes restoration more spatially focused on hallucination-prone regions. For instance, as shown in the red boxes in the second row (third and fourth columns) of Fig.~\ref{fig:ablation_experiments}, the intensities of white matter, which is the high-risk region, undergo more pronounced correction. Such focused correction brings white matter intensity closer to the ground truth and makes the gray--white matter contrast more distinct. 
Within the enhancer, repeated loop reasoning on high-risk regions makes previously ambiguous anatomical structures much clearer, especially in the red zoom-ins in the third row (fourth and fifth columns) of Fig.~\ref{fig:ablation_experiments}. Finally, fine-tuning the decoder yields the best reconstruction.

The quantitative results in Table~\ref{tab:64mt_to_3t_ablation} further confirm the effectiveness of these designs on the 64mT-to-3T task. Specifically, with five sampling steps, compared with the model without plans, atlas-based masks, loop reasoning, or decoder fine-tuning, the full model improves PSNR by 4.03 dB and SSIM by 12.6 percentage points averaged across modalities. Likewise, Table~\ref{tab:3t_to_7t_ablation} shows gains of 3.88 dB in PSNR and 15.9 percentage points in SSIM on the 3T-to-7T task under the same five-step setting. The effect of loop times is non-monotonic: performance generally improves when the number of loops increases from zero (no loop reasoning) to a small value, but degrades when more loops are used. This shows that additional loops yield diminishing refinement gains, while errors, including positional errors, accumulate with excessive loops and gradually dominate.

\begin{table}[!tb]
  \centering
  \setlength{\tabcolsep}{1.6pt}
  \caption{\textbf{Downstream results of UniField++.} Metrics are shown for input / virtual high-field MRI, using SynthSeg on real high-field MRI as the reference. $n_{\mathrm{test}}$ is the number of test volumes per modality. E denotes MRI enhancement.}
  \label{tab:downstream_segmentation_combined}
  \fontsize{8}{9.5}\selectfont
  \renewcommand{\arraystretch}{1.0}
  \resizebox{\columnwidth}{!}{%
  \begin{NiceTabular}{@{}cccc*{4}{c}@{}}
    \toprule
    \Block{2-1}{Task} & \Block{2-1}{Modality} & \Block{2-1}{\(n_{\mathrm{test}}\)} & \Block{2-1}{Setting}
      & Macro & Foreground & Macro & Volume \\
      & & & & Dice \(\uparrow\) & Dice \(\uparrow\) & IoU \(\uparrow\) & Similarity \(\uparrow\) \\
    \specialrule{\lightrulewidth}{0.8pt}{0.8pt}
    \Block{6-1}{64mT $\rightarrow$ 3T} & \Block{2-1}{T1} & \Block{2-1}{17} & w/o E & 0.5366 & 0.8573 & 0.3921 & 0.9315 \\
    & & & with E & \textbf{0.7803} & \textbf{0.9205} & \textbf{0.6538} & \textbf{0.9723} \\
     & \Block{2-1}{T2} & \Block{2-1}{17} & w/o E & 0.7220 & 0.9082 & 0.5788 & 0.9617 \\
    & & & with E & \textbf{0.8350} & \textbf{0.9404} & \textbf{0.7226} & \textbf{0.9874} \\
     & \Block{2-1}{FLAIR} & \Block{2-1}{12} & w/o E & 0.4396 & 0.8343 & 0.3127 & 0.8925 \\
    & & & with E & \textbf{0.7793} & \textbf{0.9317} & \textbf{0.6486} & \textbf{0.9828} \\
    \specialrule{\lightrulewidth}{0.8pt}{0.8pt}
    \Block{4-1}{3T $\rightarrow$ 7T} & \Block{2-1}{T1} & \Block{2-1}{6} & w/o E & 0.8294 & 0.9420 & 0.7169 & 0.9852 \\
    & & & with E & \textbf{0.8412} & \textbf{0.9468} & \textbf{0.7333} & \textbf{0.9898} \\
     & \Block{2-1}{T2} & \Block{2-1}{6} & w/o E & 0.7747 & 0.9302 & 0.6446 & 0.9723 \\
    & & & with E & \textbf{0.7950} & \textbf{0.9359} & \textbf{0.6710} & \textbf{0.9752} \\
    \specialrule{\heavyrulewidth}{0.8pt}{0.8pt}
  \end{NiceTabular}%
  }
\vspace{-2pt}
\end{table}

\begin{figure}[!tb]
    \centering
    \includegraphics[width=1.00\linewidth]{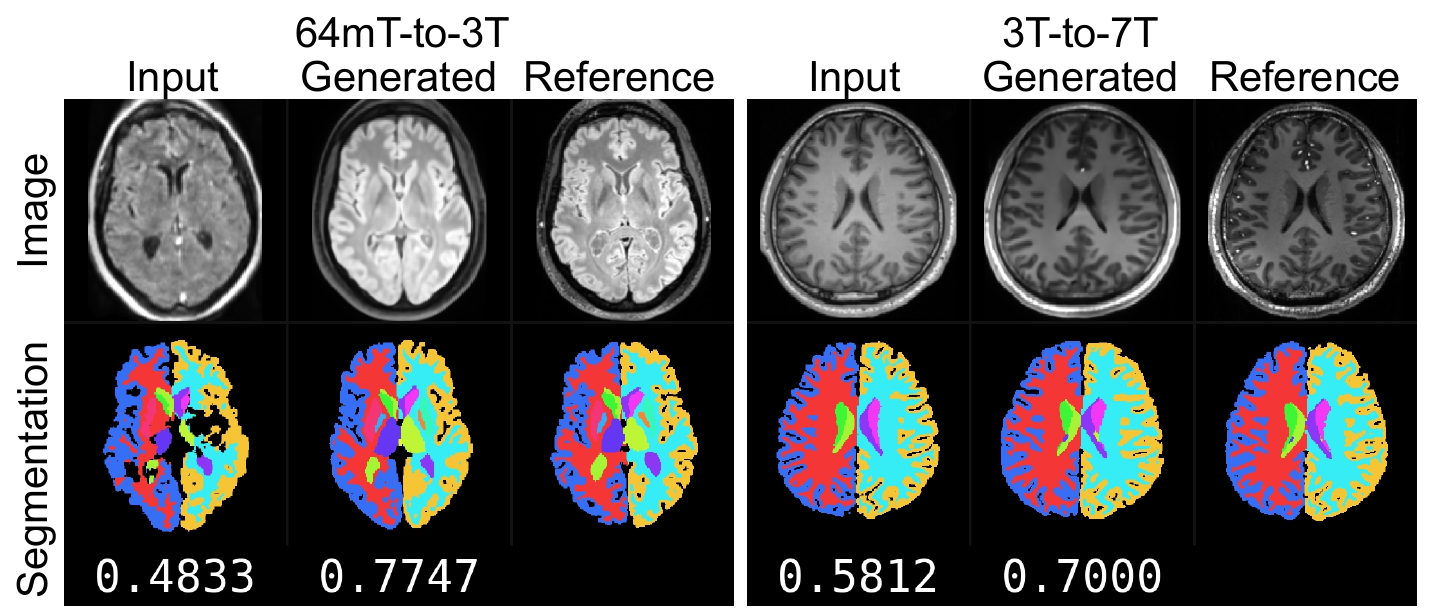}
    \caption{\textbf{Qualitative downstream segmentation results.} The numbers below the segmentation maps are Dice scores.}
    \label{fig:downstream_results}

\end{figure}

\subsection{Downstream Segmentation Validation}

To verify the downstream applicability of our method, following the evaluation approach of Gopinath et al. \cite{Gopinath2026LowField}, we assess whether field-strength enhancement improves the accuracy of whole-brain segmentation. In the downstream task, we perform segmentation with 31 foreground anatomical labels, including the brainstem, hippocampus, and amygdala, on the internal test data listed in Table~\ref{Dataset}. Specifically, SynthSeg v1 \cite{Billot2023SynthSeg} is independently applied to the original low-field MRI, our virtual high-field MRI, and the paired real high-field MRI, with the segmentation obtained from the real high-field MRI serving as the reference segmentation.

As shown in Table~\ref{tab:downstream_segmentation_combined}, segmentation based on our virtual high-field images achieves substantially higher agreement with the high-field reference than segmentation directly performed on the original low-field images. Notably, the improvement is particularly pronounced for the 64mT-to-3T task, where the average Macro Dice across T1, T2, and FLAIR increases from 0.5661 on the original 64mT MRI to 0.7982 on the synthesized 3T MRI, corresponding to an absolute improvement of 0.2321 and a relative improvement of approximately 41.0\%. 

Qualitatively, as shown in Fig. \ref{fig:downstream_results}, for the 64mT-to-3T task, the low-field segmentation incorrectly assigns substantial portions of brain tissue to the background, visible as black regions within the brain. Our enhanced images effectively reduce such errors. Additionally, for the 3T-to-7T task, the segmentations obtained from the virtual 7T images are closer to the reference segmentation in many details. For instance, regarding the lateral ventricles, shown as green and pinkish-purple areas in the center of the images, the shapes and fine structures such as edge contours in the virtual 7T segmentation are more similar to those in the reference segmentation.

\section{Conclusion}
In this paper, we propose \textbf{UniField++}, a unified multi-agent framework for MRI field-strength enhancement. In the enhancement process, UniField++ first anticipates potential hallucinations based on knowledge of field strength and brain anatomy. The framework acquires this knowledge by learning from multiple field-strength enhancement tasks and radiologist annotations. After the anticipation, the VLM planner in UniField++ can identify high-risk anatomical regions and generate enhancement plans, which are then fed to our flow-matching-based generator. Then, the generator allocates its reasoning capacity according to the predicted risks, repeatedly reasoning about high-risk regions through shared loop transformer blocks according to the enhancement plans. Extensive experiments demonstrate the effectiveness of UniField++. Across the five evaluated tasks, UniField++ improves PSNR by an average of 2.74 dB and SSIM by 14.05 percentage points over the six comparison methods. Moreover, ablation studies verify the contribution of the proposed components: with five sampling steps, the full model yields average gains of 3.97 dB in PSNR and 13.92 percentage points in SSIM over the plain baseline. Additionally, evaluation on an external cohort from BTH further demonstrates the framework's excellent ability to generalize to the unseen source field strength. Furthermore, downstream whole-brain segmentation shows an average Macro Dice improvement of 0.1457 across all five task--modality combinations over the original low-field images, confirming that field-strength enhancement provides more reliable anatomical information for subsequent analysis.

\normalem
{\small
\bibliographystyle{IEEEtran}
\balance
\bibliography{egbib}
}

\end{document}